\PassOptionsToPackage{colorlinks=true, linkcolor=blue, citecolor=blue, urlcolor=blue}{hyperref}
\PassOptionsToPackage{table}{xcolor}
\RequirePackage{amsmath}

\documentclass[runningheads]{llncs}
\usepackage[T1]{fontenc}
\usepackage{lmodern}
\usepackage{microtype}
\usepackage{graphicx,verbatim}
\usepackage{orcidlink}
\usepackage{amsfonts}
\usepackage{booktabs}
\usepackage{tabularx}
\usepackage{threeparttable}
\usepackage{multirow}
\usepackage{lineno}
\usepackage{xcolor}
\usepackage{pifont}
\usepackage{ulem}
\usepackage{xspace}
\let\lncsemail\email
\renewcommand{\email}[1]{\href{mailto:#1}{\lncsemail{#1}}}

\newcommand{\cmark}{\ding{51}} 
\newcommand{\xmark}{\ding{55}}
\begin{document}

\title{Lost in the Folds: When Cross-Validation Is Not a Deep Ensemble for Uncertainty Estimation}
\titlerunning{Lost in the Folds}



\author{
Tristan Kirscher\inst{1,2,3}\orcidlink{0009-0004-6646-6548}\thanks{Corresponding author: \email{tristan.kirscher@unistra.fr}\\ \textsuperscript{\dag}These authors share last authorship.}
\and Markus Bujotzek\inst{3,4}\orcidlink{0009-0006-4550-428X}
\and Yannick Kirchhoff\inst{3,5}\orcidlink{0000-0001-8124-8435}
\and Maximilian Rokuss\inst{3,5}\orcidlink{0009-0004-4560-0760}
\and Fabian Isensee\inst{3,6}\orcidlink{0000-0002-3519-5886}
\and Kim-Celine Kahl\inst{3,5}\textsuperscript{\dag}\orcidlink{0000-0002-4061-9073}
\and Balint Kovacs\inst{3,4}\textsuperscript{\dag}\orcidlink{0000-0002-1191-0646}
\and Klaus Maier-Hein\inst{3,7}\textsuperscript{\dag}\orcidlink{0000-0002-6626-2463}}
\authorrunning{T. Kirscher et al.}
\institute{
ICube Laboratory, CNRS UMR-7357, University of Strasbourg, Strasbourg, France
\and CLCC Institut-Strauss, Strasbourg, France
\and German Cancer Research Center (DKFZ) Heidelberg, Division of Medical Image Computing, Heidelberg, Germany
\and Medical Faculty Heidelberg, Heidelberg University, Heidelberg, Germany
\and Faculty of Mathematics and Computer Science, University of Heidelberg, Germany
\and Helmholtz Imaging, German Cancer Research Center, Heidelberg, Germany
\and Pattern Analysis and Learning Group, Department of Radiation Oncology, Heidelberg University Hospital, Heidelberg, Germany
}
  
\maketitle              

\begin{abstract}
Ensemble disagreement is widely used as a proxy for epistemic uncertainty in medical image segmentation. In practice, many studies form ensembles via K-fold cross-validation (CV), yet refer to them as “deep ensembles” (DE). Because CV members are trained on different data subsets, their disagreement mixes seed-driven variability with data-exposure effects, which can change how uncertainty should be interpreted.
We audit recent segmentation uncertainty studies and find that terminology--implementation mismatches are common. We then compare a standard 5-fold CV ensemble to a 5-member DE (fixed training set, different random seeds) under otherwise identical configurations on three multi-rater segmentation datasets spanning three modalities. We evaluate uncertainty for calibration, failure detection, ambiguity modeling, and robustness under distribution shift.
DE match segmentation accuracy while improving calibration and failure detection, whereas CV ensembles sometimes correlate more strongly with inter-rater variability on the studied datasets. Thus, ensemble construction should be chosen to match the research question: DE for reliability-oriented use (e.g., selective referral/failure detection) and CV ensembles as a proxy for ambiguity. We provide a lightweight nnU-Net modification enabling DE training within the default pipeline.
\end{abstract}

\section{Introduction}

Ensemble methods are widely used in medical image segmentation for robustness and uncertainty quantification, particularly in reliability-critical clinical pipelines. Among these methods, deep ensembles (DE), i.e. multiple instances of the same model trained on the same training set with different random initializations, have emerged as a simple and effective approximation to Bayesian model averaging \cite{lakshminarayanan2017deep,ovadia2019can}. In high-capacity neural networks, deep ensembles consistently outperform alternatives such as Monte Carlo dropout in terms of calibration, robustness under dataset shift, and failure detection \cite{ovadia2019can}.

In parallel, nnU-Net has become a de facto standard framework for medical image segmentation due to its automated configuration, strong empirical performance, and broad adoption across modalities and tasks \cite{isensee2021nnunet,isensee2024nnu}. By default, nnU-Net employs five-fold cross-validation (CV) to estimate in-distribution performance and averages fold-specific predictions at inference time to maximize segmentation accuracy. Practitioners often reuse these CV ensembles for uncertainty estimation rather than retraining deep ensembles (DE).

Increasingly, disagreement among these CV models is reused as a proxy for epistemic uncertainty in downstream applications such as quality control and human-in-the-loop review \cite{jungo2020analyzing,mehrtash2021confidence,khalili2024uncertainty}. However, CV ensembles and DE differ in a fundamental and often overlooked way. In a deep ensemble, all models are trained on the same training set, and prediction variability reflects uncertainty in model parameters given the available data. In contrast, each model in a CV ensemble is trained on a different subset of the training data. As a result, disagreement within a CV ensemble conflates epistemic uncertainty with variability induced by incomplete or heterogeneous data exposure. This distinction is particularly relevant in medical imaging, where rare structures, underrepresented pathologies, or uneven fold composition can cause elevated disagreement that reflects missing training examples rather than genuine uncertainty under the full data distribution \cite{ovadia2019can}. Our goal is to investigate these ensemble formulations to determine whether they induce systematically different uncertainty behavior, and which is preferable for a given downstream task.

In practice, uncertainty estimates derived from segmentation models are often thresholded, ranked, or aggregated to trigger concrete actions, such as selective referral to human experts \cite{khalili2024uncertainty}, automated failure detection \cite{zenk2025benchmarking}, or quality control and identification of reduced model performance \cite{alves2024uncertainty,vanaalst2025uncertainty}. Therefore, it is crucial to develop an understanding of how different ensemble constructions affect these follow-up tasks.

\textbf{Contributions.} This paper clarifies the distinction between CV ensembles and DE and addresses their conflation in the literature. The contributions are:
\begin{enumerate}
    \item \textbf{Terminology audit.} We document, via a survey of uncertainty in segmentation papers, that the term deep ensemble is frequently used to describe a K-fold CV ensemble, similar to the default nnU-Net behavior, despite its members being trained on different data subsets (\autoref{tab:de_vs_cv_nnunet}). We therefore distinguish CV ensembles from DE (same data, different seeds) throughout.
    \item \textbf{Controlled comparison across downstream tasks.} Using the same network architecture and hyperparameters, we compare 5-fold CV ensembles against 5-member DE on multi-rater datasets spanning different modalities and evaluate segmentation, calibration, ambiguity modeling, failure detection, and shift robustness under a unified protocol (\autoref{tab:full_results}, Fig. \ref{fig:ace}-\ref{fig:referral}).
    \item \textbf{Practical implementation.} We provide a minor modification to the nnU-Net pipeline to support deep ensembles as originally described in \cite{lakshminarayanan2017deep}.
    \item \textbf{Task-dependent recommendations.} We show that the ensemble construction materially affects uncertainty behavior: DE provide more reliable calibration and failure detection at comparable Dice, while CV can better reflect annotation ambiguity in some settings. Consequently, we provide task-dependent recommendations for using and reporting ensemble uncertainty in reliability-oriented segmentation studies.
\end{enumerate}

\begin{table*}[htb]
\centering
\small
\setlength{\tabcolsep}{2.2pt}
\caption{Use of ensemble methods for uncertainty estimation in segmentation studies (2020--2025). Papers were identified via a literature search, followed by manual screening. SD: same training set across ensemble members. Align: method--claim alignment.}
\label{tab:de_vs_cv_nnunet}

\begin{tabular}{l l c l c c}
\toprule
\textbf{Paper} & \textbf{Task} & \textbf{Claim} & \textbf{Actual} & \textbf{SD} & \textbf{Align} \\
\midrule

Jungo et al.~\cite{jungo2020analyzing} & Brain MRI & DE & 10-fold CV & \xmark & \xmark \\
Alves et al.~\cite{alves2024uncertainty} & Multi-organ CT & DE & 5-fold CV & \xmark & \xmark \\
van Aalst et al.~\cite{vanaalst2025uncertainty} & Head \& Neck OAR CT & DE & 5-fold CV & \xmark & \xmark \\
Khalili et al.~\cite{khalili2024uncertainty} & CAMELYON16 (WSI) & DE & 5-fold CV & \xmark & \xmark \\
Gotkowski et al.~\cite{gotkowski2022i3deep} & Multi-dataset (CT/MRI) & DE & 5-fold CV & \xmark & \xmark \\
Schwab et al.~\cite{schwab2025disagreement} & Cardiac MRI (LGE) & DE & 5-fold CV & \xmark & \xmark \\
Rosas-Gonzalez et al.~\cite{rosasgonzalez2021ae_aunet} & Brain MRI & DE & 5-fold CV & \xmark & \xmark \\

G\"ottlich et al.~\cite{gottlich2023ai} & Multi-organ CT & CV & 5-fold CV & \xmark & \cmark \\
Gade et al.~\cite{gade2024conformal} & Prostate MRI & CV & 5-fold CV & \xmark & \cmark \\
Kucyba{\l}a et al.~\cite{kucybala2024uq} & Multi-dataset (CT/MRI) & CV & 5-fold CV & \xmark & \cmark \\

Molchanova et al.~\cite{molchanova2025explaining} & Multiple sclerosis MRI & DE & DE & \cmark & \cmark \\
Zhao et al.~\cite{zhao2022bayesian} & Cardiac MRI & DE & DE & \cmark & \cmark \\
Zenk et al.~\cite{zenk2025benchmarking} & Multi-dataset (CT/MRI) & DE & DE & \cmark & \cmark \\
Buddenkotte et al.~\cite{buddenkotte2023calibrating} & Ovarian/Kidney CT & Ens. & Diff. losses & \cmark & \cmark \\

\bottomrule
\end{tabular}
\end{table*}

\section{Related Work}

\paragraph{\textbf{Deep ensembles and epistemic uncertainty.}}
Deep ensembles were introduced as a scalable and effective approach to epistemic uncertainty estimation by training multiple networks on the same dataset with different random initializations and stochastic optimization trajectories \cite{lakshminarayanan2017deep}. Despite their simplicity, deep ensembles have been shown to provide strong uncertainty estimates, often outperforming approximate Bayesian methods such as Monte Carlo dropout in terms of calibration, robustness, and out-of-distribution behavior \cite{ovadia2019can,kendall2017uncertainties,kahl2024values}. 

\paragraph{\textbf{Uncertainty estimation in medical image segmentation.}}
Uncertainty quantification has been widely studied in medical image segmentation, motivated by applications such as quality control, error detection, and human-in-the-loop decision support \cite{jungo2020analyzing,mehrtash2021confidence}. Ensemble-based methods have consistently demonstrated improved calibration and error awareness compared to single-model predictions~\cite{mehrtash2021confidence,huang2023review,kahl2024values}. 
Recently, there has been an increased emphasis on systematic validation of uncertainty estimates with respect to concrete downstream applications. This is formalized in the ValUES framework by identifying calibration, failure detection, and ambiguity modeling as key uncertainty evaluation scenarios in semantic segmentation \cite{kahl2024values}. For calibration, spatially resolved evaluation metrics, such as BA-ECE have been proposed to better capture the alignment between uncertainty maps and segmentation errors \cite{zeevi2025spatially}. Failure detection, on the other hand, is a task that is tackled at the subject-level, motivated by the need to flag whole subjects as failures in clinical deployment \cite{jungo2019assessing,jungo2020analyzing,zenk2025benchmarking}. In ambiguity modeling, the goal is to capture data ambiguity, for example when raters provide different segmentation masks for the same image.

\paragraph{\textbf{Ensembles in nnU-Net pipelines.}}
By default, the widely adopted state-of-the-art segmentation framework, nnU-Net, trains models in a 5-fold CV in order to estimate in-distribution performance and then re-uses these models as a CV ensemble to boost test set performance \cite{isensee2021nnunet}. However, as shown in \autoref{tab:de_vs_cv_nnunet}, several recent studies implicitly or explicitly interpret disagreement among nnU-Net CV models as epistemic uncertainty, often referring to these as DE \cite{jungo2020analyzing,alves2024uncertainty,vanaalst2025uncertainty,khalili2024uncertainty,gotkowski2022i3deep,schwab2025disagreement,rosasgonzalez2021ae_aunet}. Only a limited number of works explicitly adhere to the DE definition by training all ensemble members on identical data with independent random seeds \cite{zhao2022bayesian,molchanova2025explaining,zenk2025benchmarking}.

\section{Problem Statement}
We study supervised medical image segmentation with dataset 
$\mathcal{D} = \{(x_i, y_i)\}_{i=1}^{N}$, where $x_i$ is an image and $y_i$ its mask. 
A segmentation network $f(x;\theta)$ predicts $p(y \mid x, \theta)$.
We compare two ensemble constructions for uncertainty estimation:

\paragraph{\textbf{Deep Ensembles (DE).}}
A deep ensemble consists of $M$ independently trained models $\{f(x;\theta_m)\}_{m=1}^{M}$, where each parameter vector $\theta_m$ is obtained by optimizing the same objective on the same training dataset $\mathcal{D}$, but with different random initializations and stochastic training~\cite{lakshminarayanan2017deep}. Formally,
\begin{equation}
\theta_m \sim P(\theta \mid \mathcal{D}), \quad m = 1,\dots,M,
\end{equation}
where the randomness induced by initialization, data shuffling, and augmentation implicitly samples multiple modes of the weight posterior. The ensemble predictive distribution for a new input $x$ is given by
$p(y \mid x, \mathcal{D}) \approx \frac{1}{M} \sum_{m=1}^{M} p(y \mid x, \theta_m)\,$,
and the variability among predictions $\{f(x;\theta_m)\}$ reflects the epistemic uncertainty of the model given the full dataset $\mathcal{D}$.

\paragraph{\textbf{Cross-Validation (CV) Ensembles.}}
In a $K$-fold CV ensemble, the dataset is partitioned into disjoint folds $\{\mathcal{D}_k^{\text{val}}\}_{k=1}^{K}$, with corresponding training subsets $\mathcal{D}_k^{\text{train}} = \mathcal{D} \setminus \mathcal{D}_k^{\text{val}}$. Each model $f(x;\theta_k)$ is trained on a different data subset:
\begin{equation}
\theta_k \sim P(\theta \mid \mathcal{D}_k^{\text{train}}), \quad k = 1,\dots,K\,.
\end{equation}
The ensemble predictive distribution is 
$p(y \mid x) \approx \frac{1}{K} \sum_{k=1}^{K} p(y \mid x, \theta_k)\,$,
representing a mixture of models conditioned on different training sets. Disagreement between these models arises not only from posterior uncertainty, but also from the fact that each $\theta_k$ is optimized with incomplete data $\mathcal{D}_k^{\text{train}}$ (missing $\mathcal{D}_k^{\text{val}}$).

\paragraph{\textbf{Key Distinction.}}
DE aim to approximate Bayesian model averaging under a fixed dataset. 
In contrast, CV ensembles aggregate models trained on different and incomplete subsets of the data. 
Consequently, their disagreement reflects both epistemic uncertainty and variability induced by data subsampling. 
\textit{A model may therefore appear uncertain due to limited exposure to specific training examples, rather than uncertainty under the full-data distribution.}

\section{Experimental Setup}
The experiments aim to isolate the effect of ensemble construction on uncertainty estimation. Specifically, we compare CV ensembles, as commonly used in nnU-Net-based studies, with DE that adhere to the original definition from~\cite{lakshminarayanan2017deep}.

\paragraph{\textbf{Datasets.}}
Ensemble uncertainty is evaluated on multi-rater medical image segmentation datasets spanning 2D and 3D modalities, including MRI, CT, and retinal fundus imaging (cf.~\autoref{tab:datasets}). 
The datasets cover three types of consensus: manual expert delineations, STAPLE~\cite{staple}, and majority voting, thereby reflecting different annotation aggregation paradigms. 
All tasks provide independent rater annotations, enabling ambiguity analysis. 
Training uses all available rater annotations by treating each (image, rater) pair as a separate training example. To prevent leakage - i.e., the same image appearing in both training and validation with different rater labels - we generate splits with a K-fold procedure while grouping by image identity.
Preprocessing follows the standard nnU-Net pipeline without task-specific modifications, isolating ensemble construction as the primary variable.
For out-of-distribution (OOD) evaluation, dataset-specific splits were defined to induce clinically meaningful shifts.
For CURVAS, Group~C (cysts with contour-altering pathologies; $n{=}23$) is held out as OOD. 
For RIGA, images from the Magrabi Eye Center ($n{=}95$) were held out, representing a shift in acquisition center and population.
For GoldAtlas, the patients from acquisition site~3 ($n{=}4$) were treated as OOD. 

\begin{table}[t]
\small
\centering
\caption{Overview of heterogeneous multirater-datasets used in this study, covering various medical modalities and different methods for generating consensus masks.}
\label{tab:datasets}
\begin{tabularx}{\textwidth}{l X c c c c}
\toprule
\textbf{Dataset} & \textbf{Modality} & \textbf{Samples} & \textbf{Classes} & \textbf{Raters} & \textbf{Consensus} \\
\midrule
GoldAtlas~\cite{nyholm2018goldatlas} 
& T2w MRI (3D) 
& 19 
& 9 
& 5 
& Manual (experts) \\


CURVAS~\cite{riera2024curvas} 
& CT (3D) 
& $89^{*}$
& 4 
& 3 
& STAPLE \\

RIGA~\cite{almazroa2018riga} 
& Fundus RGB (2D) 
& 749 
& 2 
& 6 
& Majority vote\\
\bottomrule
\end{tabularx}
\footnotesize{$^{*}$One case was excluded due to annotation spacing inconsistencies from one rater.}
\end{table}

\paragraph{\textbf{Network and Training Configuration.}}
All experiments use nnU-Net (v2.4.1) full-resolution configuration with the \texttt{ResEncM} preset and default hyperparameters \cite{isensee2021nnunet}. Network architecture, training schedule, and data augmentation are automatically determined by the nnU-Net framework for each task. Training is run for a fixed number of epochs with a fixed learning-rate schedule (no early stopping and no validation-based learning-rate adaptation). The validation split is used only for reporting, not for model selection; inference uses the final checkpoint for all models. This protocol is identical for CV and DE members, so differences reflect the ensemble construction (shared vs varying training subsets) rather than checkpointing or optimization artifacts. 

\paragraph{\textbf{Ensemble Configurations.}}
We define two ensemble configurations using the same pool of training cases.
For the \textbf{CV-Ensemble}, 5-fold CV is performed using nnU-Net’s default data splits. Each model is trained on $80\%$ of the data and validated on the remaining fold. For the \textbf{DE}, $M=5$ models are trained independently on all available training data, differing only in random initialization and stochastic training effects (data shuffling, augmentation, and optimization). This configuration satisfies the defining criteria of a deep ensemble \cite{lakshminarayanan2017deep}.
Apart from data exposure, all training and inference settings are identical across ensembles.

\paragraph{\textbf{Uncertainty Quantification and Evaluation.}}
During inference, voxel-wise softmax probability maps $p_m$ are computed for each ensemble member $m$.
Ensemble predictions are obtained by averaging class probabilities across models, $\bar{p}=\frac{1}{M}\sum_{m=1}^{M} p_m$, and the mean hard prediction is given by $\bar{y}(v)=\arg\max_c \bar{p}_c(v)$.
To assess segmentation performance, we compute the Dice score (DSC) between the mean prediction $\bar{y}$ and the consensus segmentation $\bar{y}^*$, i.e., $\mathrm{DSC}(\bar{y},\bar{y}^*)$.
As calibration metrics, the Average Calibration Error (ACE) \cite{jungo2020analyzing,kahl2024values} and the Boundary-Aware Expected Calibration Error (BA-ECE) \cite{zeevi2025spatially} are calculated on a voxel level, where the confidence is defined as $\mathrm{conf}(v)=\max_c \bar{p}_c(v)$. These confidences are then grouped into bins, and the accuracy in a bin $B_m$ is calculated as $acc(B_m)=\frac{1}{|B_m|N}\sum_{v\in B_m}\sum_{n=1}^N[\,\bar{y}(v)=y_n^*(v)\,]$.
Ambiguity modeling is evaluated using the Normalized Cross-Correlation (NCC) between the predictive entropy map $H(\bar{p})$ and the rater variance map \cite{kahl2024values}, and the Generalized Energy Distance (GED) between the individual ensemble predictions $\{y_m\}_{m=1}^{M}$ and the individual rater annotations $\{y_n^*\}_{n=1}^{N}$ \cite{kahl2024values}.
Failure detection is performed at a sample level following \cite{zenk2025benchmarking}, where cases are ranked by inter-model disagreement, $u=\mathbb{E}_{i\neq j}\!\left[1-\mathrm{DSC}(y_i,y_j)\right]$, and the risk is defined as $r=1-\mathrm{DSC}(\bar{y},\bar{y}^*)$.
Non-parametric bootstrap confidence intervals (CI) are used to compare uncertainty and agreement measures between CV ensembles and DE (see \autoref{tab:full_results}).

\section{Results}

\begin{figure}[tb]
    \centering
    \includegraphics[width=.92\linewidth]{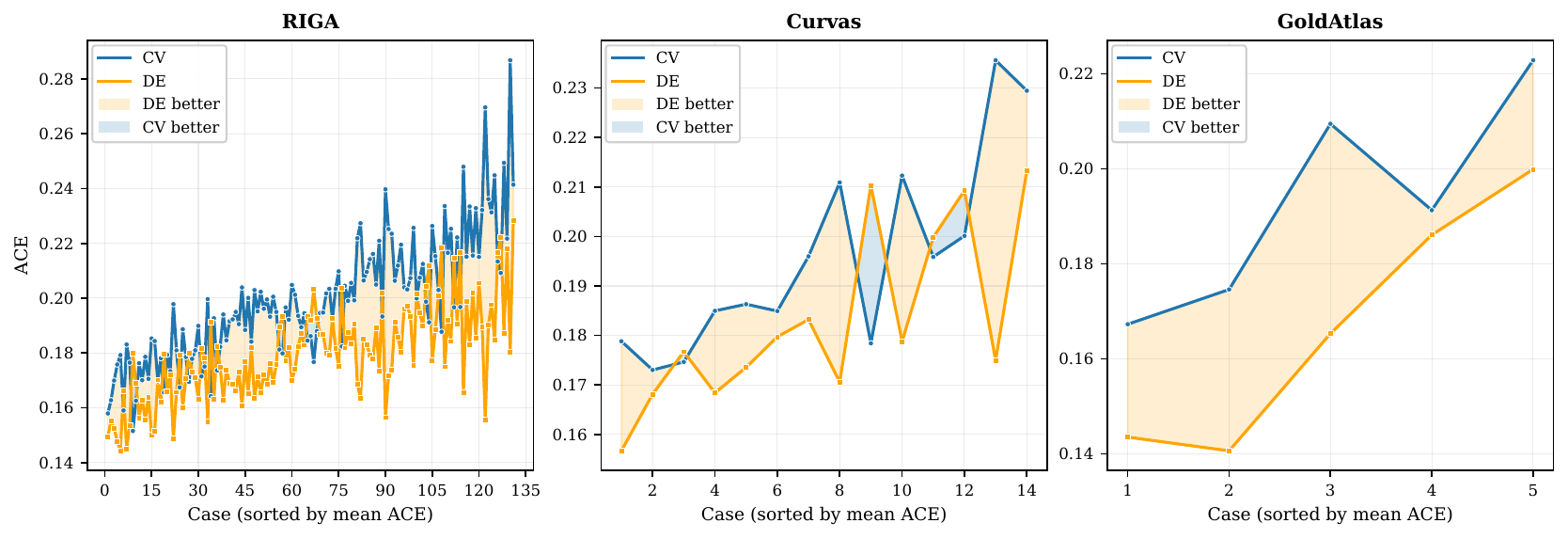}
    \caption{Per-case Average Calibration Error (ACE) on the in-distribution test set for Cross-Validation (CV) and Deep Ensemble (DE) methods across all datasets.}
    \label{fig:ace}
\end{figure}

\begin{figure}[t]
    \centering
    \includegraphics[width=.9\linewidth]{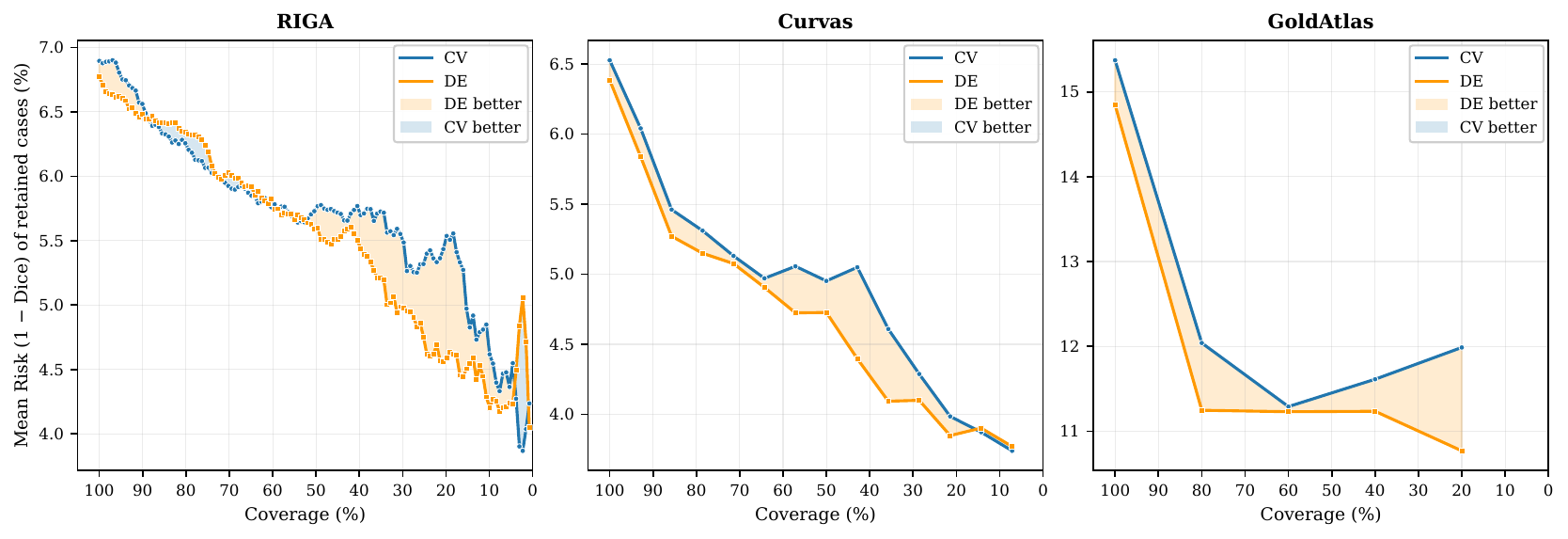}
    \caption{Referral curves showing the mean risk (1$-$Dice) of in-distribution retained cases as a function of coverage for CV and DE methods across all datasets.}
    \label{fig:referral}
\end{figure}

\autoref{tab:full_results} summarizes performance and uncertainty across datasets, while Figs.~\ref{fig:ace}--\ref{fig:referral} visualize calibration and failure detection behavior, respectively.
Segmentation accuracy remains essentially unchanged: DE achieve DSC comparable to CV ensembles across all datasets.
In contrast, uncertainty properties differ systematically. DE consistently improve voxel-wise calibration, reducing ACE across datasets (\autoref{fig:ace}) and yielding lower BA-ECE.
At the sample level, DE provide stronger failure detection, reflected by lower AURC and reduced risk at matched coverage in referral curves (\autoref{fig:referral}).
Interestingly, CV ensembles better align with inter-rater ambiguity on CURVAS and RIGA (higher NCC and/or lower GED), suggesting that fold-induced data variability can partially mimic annotation ambiguity.
Under distribution shift, differences between CV and DE are generally modest. While DE show slight advantages in some settings, neither ensemble type consistently dominates across datasets, indicating comparable robustness under the considered OOD splits. Overall, these findings indicate that the appropriate ensemble construction depends on the downstream task: calibrated epistemic uncertainty for reliability-oriented decision-making favors DE, whereas ambiguity modeling may benefit from CV ensembles (\autoref{tab:full_results}).

\begin{table*}[tb]
\centering
\caption{Comparison of CV vs DE performance.
ID rows (CV, DE) report mean and 95\% percentile bootstrap CIs over test cases.
Paired bootstrap ($B{=}10{,}000$) is performed on per-case differences (DE$-$CV); for AURC, curves are recomputed per resample.
The OOD row reports only the paired OOD $\Delta$ effect size as DE improvement $\Delta=(DE{-}CV)\times s$ with CI, where $s{=}{+}1$ if higher is better and $s{=}{-}1$ if lower is better.
Stars indicate significance of the paired DE$-$CV comparison within the corresponding row ($^{*}p<0.05$, $^{\dagger}p<0.001$). Within each dataset, the better method (per metric direction) is \textbf{bolded}. Gray rows indicate the overall better-performing method according to this metric.
SEG: Segmentation; CAL: Calibration; AM: Ambiguity Modeling; FD: Failure Detection.
\textbf{All reported values are multiplied by 100 for compactness.}}
\label{tab:full_results}
\setlength{\tabcolsep}{3pt}
\begin{tabularx}{\textwidth}{l|ll|XXX}
\toprule
Task & Metric & Setting & GoldAtlas & CURVAS & RIGA \\
\midrule

\multirow{3}{*}{SEG}
& \multirow{3}{*}{DSC $\uparrow$}
& CV
& 84.6 {\footnotesize (74.6, 88.4)}
& 93.5 {\footnotesize (91.5, 94.6)}
& 93.1 {\footnotesize (92.6, 93.6)} \\
&  & \cellcolor{gray!20}DE
& \cellcolor{gray!20}\textbf{85.2} {\footnotesize (74.3, 88.9)}
& \cellcolor{gray!20}\textbf{93.6} {\footnotesize (91.4, 94.8)}
& \cellcolor{gray!20}\textbf{93.2}$^{\dagger}$ {\footnotesize (92.7, 93.7)} \\
&  & \footnotesize OOD $\Delta$
& \footnotesize +3.3 (-0.3, 8.8)
& \footnotesize +0.4$^{*}$ (0.1, 0.7)
& \footnotesize +0.1 (-0.6, 0.8) \\

\cmidrule(lr){1-6}

\multirow{6}{*}{CAL}
& \multirow{3}{*}{ACE $\downarrow$}
& CV
& 19.3 {\footnotesize (17.5, 21.1)}
& 19.6 {\footnotesize (18.7, 20.8)}
& 19.9 {\footnotesize (19.5, 20.3)} \\
&  & \cellcolor{gray!20}DE
& \cellcolor{gray!20}\textbf{16.7}$^{\dagger}$ {\footnotesize (14.7, 18.7)}
& \cellcolor{gray!20}\textbf{18.3}$^{*}$ {\footnotesize (17.5, 19.3)}
& \cellcolor{gray!20}\textbf{17.9}$^{\dagger}$ {\footnotesize (17.7, 18.2)} \\
&  & \footnotesize OOD $\Delta$
& \footnotesize +2.5 (-0.6, 6.0)
& \footnotesize +1.5$^{\dagger}$ (0.8, 2.2)
& \footnotesize +0.8 (-0.4, 2.0) \\

\cmidrule(lr){2-6}
& \multirow{3}{*}{BA-ECE $\downarrow$}
& CV
& 7.1 {\footnotesize (5.7, 8.8)}
& 6.8 {\footnotesize (6.1, 7.8)}
& 21.5 {\footnotesize (20.9, 22.0)} \\
&  & \cellcolor{gray!20}DE
& \cellcolor{gray!20}\textbf{6.4}$^{\dagger}$ {\footnotesize (5.2, 8.5)}
& \cellcolor{gray!20}\textbf{6.4}$^{\dagger}$ {\footnotesize (5.8, 7.4)}
& \cellcolor{gray!20}\textbf{20.6}$^{\dagger}$ {\footnotesize (20.1, 21.2)} \\
&  & \footnotesize OOD $\Delta$
& \footnotesize +1.6 (0.6, 3.2)
& \footnotesize +0.4$^{\dagger}$ (0.3, 0.4)
& \footnotesize +0.7$^{\dagger}$ (0.4, 1.1) \\

\cmidrule(lr){1-6}

\multirow{6}{*}{AM}
& \multirow{3}{*}{NCC $\uparrow$}
& \cellcolor{gray!20}CV
& \cellcolor{gray!20} 54.4 {\footnotesize (51.6, 57.0)}
& \cellcolor{gray!20}\textbf{50.3}$^{\dagger}$ {\footnotesize (48.3, 53.1)}
& \cellcolor{gray!20}\textbf{73.7}$^{\dagger}$ {\footnotesize (72.5, 74.6)} \\
&  & DE
& \textbf{54.5} {\footnotesize (51.1, 57.6)}
& 49.2 {\footnotesize (47.3, 51.8)}
& 72.9 {\footnotesize (71.8, 73.8)} \\
&  & \footnotesize OOD $\Delta$
& \footnotesize +1.4 (-0.6, 3.3)
& \footnotesize -0.7$^{\dagger}$ (-0.9, -0.5)
& \footnotesize -0.2 (-0.9, 0.5) \\

\cmidrule(lr){2-6}
& \multirow{3}{*}{GED $\downarrow$}
& \cellcolor{gray!20}CV
& \cellcolor{gray!20}\textbf{16.2} {\footnotesize (12.9, 21.2)}
& \cellcolor{gray!20}\textbf{8.8} {\footnotesize (7.2, 11.3)}
& \cellcolor{gray!20}\textbf{8.2}$^{\dagger}$ {\footnotesize (7.7, 8.9)} \\
&  & DE
& 16.8 {\footnotesize (11.8, 29.9)}
& 9.2 {\footnotesize (7.5, 12.1)}
& 8.6 {\footnotesize (8.1, 9.3)} \\
&  & \footnotesize OOD $\Delta$
& \footnotesize +2.7 (0.4, 6.0)
& \footnotesize -0.1 (-0.4, 0.3)
& \footnotesize -0.9 (-2.6, 0.7) \\

\cmidrule(lr){1-6}

\multirow{3}{*}{\shortstack{FD}}
& \multirow{3}{*}{AURC $\downarrow$}
& CV
& 9.7 {\footnotesize (8.9, 13.8)}
& 4.6 {\footnotesize (3.9, 5.4)}
& 5.7 {\footnotesize (5.2, 6.2)} \\
&  & \cellcolor{gray!20}DE
& \cellcolor{gray!20}\textbf{9.3} {\footnotesize (8.7, 13.9)}
& \cellcolor{gray!20}\textbf{4.4} {\footnotesize (3.8, 5.2)}
& \cellcolor{gray!20}\textbf{5.5} {\footnotesize (5.1, 5.9)} \\
&  & \footnotesize OOD $\Delta$
& \footnotesize +2.6 (-0.7, 8.1)
& \footnotesize +0.2 (0.0, 0.6)
& \footnotesize -4.3 (-11.7, 0.3) \\

\bottomrule
\end{tabularx}
\end{table*}

\section{Discussion and Conclusion}

\begin{table}[htb]
\centering
\setlength{\tabcolsep}{10pt}
\small
\caption{Our ensemble construction recommendations for different uncertainty tasks.}
\label{tab:ensemble_task_summary}
\begin{tabular}{lcc}
\toprule
\textbf{Task} & \textbf{Deep Ensemble} & \textbf{CV Ensemble} \\
\midrule
Calibration & \cmark &  \\
Ambiguity modeling &  & \cmark \\
Distribution shift robustness & \cmark & \cmark \\
Failure detection & \cmark &  \\
\bottomrule
\end{tabular}
\end{table}

CV ensembles are often used for uncertainty estimation, not least in nnU-Net pipelines, because they are readily available and improve segmentation accuracy. Yet, our analysis shows that they are not interchangeable with DE for uncertainty estimation: since fold members are trained on different data subsets, their disagreement reflects both epistemic uncertainty and sensitivity to data exposure. This distinction is critical in reliability-oriented workflows that threshold or rank uncertainty for referral or failure detection. Empirically, DE achieve comparable segmentation performance while consistently improving calibration and failure detection. In contrast, CV ensembles better match inter-rater ambiguity on some datasets, suggesting that fold-induced data variability can mimic annotation uncertainty. Ensemble construction should therefore align with the intended downstream task: reliability-sensitive decision-making favors DE, whereas ambiguity modeling may benefit from CV ensembles.
In practice, DE also incur extra cost, as members must be trained in addition to the CV models typically produced during parameter optimization.
We recommend that future studies (i) explicitly report the ensemble construction used for uncertainty estimation and (ii) select DE or CV ensembles based on the application.
To facilitate this, we release a minimal nnU-Net change that enables deep ensembles support as described in \cite{lakshminarayanan2017deep}.
\autoref{tab:ensemble_task_summary} summarizes our task-dependent recommendations.

\section*{Code and Data Availability}
Code is available at \url{https://github.com/Kirscher/LostInFolds}. The study uses public datasets; access and reuse are subject to the original dataset providers' terms for GoldAtlas~\cite{nyholm2018goldatlas}, CURVAS~\cite{riera2024curvas}, and RIGA~\cite{almazroa2018riga}. Preprocessed data, model checkpoints, and derived predictions are not redistributed within the repository.

\section*{Acknowledgments}
This work of the Interdisciplinary Thematic Institute HealthTech, as part of the ITI 2021-2028 program of the University of Strasbourg, CNRS and Inserm, was partially supported by IdEx Unistra (ANR-10-IDEX-0002) and SFRI (STRAT'US project, ANR-20-SFRI-0012) under the framework of the French Investments for the Future Program. T.K. research visit to the Medical Image Computing team at DKFZ was supported by a CNRS GdR IASIS grant and Erasmus funding.
The present contribution is supported by the Helmholtz Association under the joint research program ``HIDSS4Health - Helmholtz Information and Data Science School for Health''.

\section*{Disclosure of Interests}
The authors have no competing interests to declare
that are relevant to the content of this article.

\newpage


\bibliographystyle{splncs04}
\bibliography{biblio}

\end{document}